\documentclass{article} 
\usepackage{iclr2025_conference,times}
\iclrfinalcopy  

\usepackage{wrapfig}

\usepackage{amsmath,amsfonts,bm}

\def\eqref#1{equation~\ref{#1}}

\def\1{\bm{1}}

\DeclareMathAlphabet{\mathsfit}{\encodingdefault}{\sfdefault}{m}{sl}
\SetMathAlphabet{\mathsfit}{bold}{\encodingdefault}{\sfdefault}{bx}{n}

\newcommand{\E}{\mathbb{E}}

\usepackage{mathtools} 
\usepackage{booktabs} 
\usepackage{tikz} 
\usepackage{caption}
\usepackage{hyperref}
\usepackage{url}
\usepackage{amssymb}
\usepackage{amsmath,mathrsfs,dsfont}
\usepackage{nicefrac}
\usepackage[algo2e]{algorithm2e}
\usepackage{algorithmic}
\usepackage{algorithm}
\usepackage{graphicx}
\usepackage{multirow}
\usepackage{caption}
\usepackage{booktabs}
\usepackage{subcaption}
\usepackage{color}
\usepackage{enumitem}
\usepackage{multirow}
\usepackage{array}
\usepackage{graphicx,tikz}
\usepackage[mathscr]{euscript}
\usepackage{amsmath,bm}
\usepackage{amsthm}
\usepackage{array}
\usepackage{graphicx,tikz}
\usepackage[mathscr]{euscript}
\usepackage{amsmath,bm}
\usepackage{amsthm}

\usepackage{makecell}

\usepackage{bm}
\usepackage{bbm}
\usepackage{color}
\newcounter{optproblem}

\newtheoremstyle{mytheoremstyle} 
    {\topsep}                    
    {\topsep}                    
    {\normalfont}                
    {}                           
    {\bfseries}                   
    {.}                          
    {.5em}                       
    {}  

\theoremstyle{mytheoremstyle}
\newtheorem{theorem}{Theorem}[section]

\newtheorem*{theorem*}{Theorem}
\newtheorem*{lemma*}{Lemma}
\newtheorem*{remark*}{Remark}

\theoremstyle{mytheoremstyle}

\theoremstyle{remark}

\DeclareMathAlphabet{\pazocal}{OMS}{zplm}{m}{n}
\DeclareMathAlphabet{\mathpzc}{OMS}{pzc}{m}{it}

\renewcommand{\hat}{\widehat}

\newcommand{\bfm}[1]{\ensuremath{\mathbf{#1}}}
\newcommand{\bfsym}[1]{\ensuremath{\boldsymbol{#1}}}

   \def\bA{\bfm A}  
     
   \def\bC{\bfm C}

  \def\bF{\bfm F}

   \def\bI{\bfm I}  
     
   \def\bK{\bfm K}  
     
\def\bm{\bfm m}

   \def\bQ{\bfm Q}  
     \def\RR{\mathbb{R}}

   \def\bU{\bfm U}

\def\bx{\bfm x}   \def\bX{\bfm X}  
   \def\bY{\bfm Y}  
     
\def\bzero{\bfm 0}

 \def\cB{{\cal  B}}
 
 \def\cD{{\cal  D}}

 \def\cW{{\cal  W}}
 \def\cX{{\cal  X}}

\def\bLambda {\bfsym {\Lambda}}

\def\hlambda{\hat{\lambda}}

\def\+#1{\mathcal{#1}}
\def\-#1{\textup{#1}}

\def\set#1{\left\{ #1 \right\}}
\def\pth#1{\left( #1 \right)}
\def\bth#1{\left[ #1 \right]}
\def\abth#1{\left | #1 \right |}

\def\defeq {\coloneqq}

\newcommand{\La}{\left\langle\kern-0.64ex\left\langle}
\newcommand{\Ra}{\right\rangle\kern-0.64ex\right\rangle}

\def\Norm#1#2{{\left\vert\kern-0.4ex\left\vert\kern-0.4ex\left\vert #1
    \right\vert\kern-0.4ex\right\vert\kern-0.4ex\right\vert}_{#2}}
\def\norm#1#2{{\left\|#1\right\|}_{#2}}

\def\ltwonorm#1{\norm{#1}{2}}
\def\fnorm#1{\norm{#1}{\textup{F}}}

\newcommand{\rank}{\textup{rank}}

\def\tr#1{\textup{tr}\left(#1\right)}

\def\indict#1{{\rm 1}\kern-0.25em{\rm I}_{\set{#1}}}

\def\set#1{\left\{#1\right\}}

\def \E {\mathbb{E}}
\def\Expect#1#2{\E_{#1}\left[#2\right]}

\def \lsim {\lesssim}

\newcommand{\beq}{\begin{equation}}
\newcommand{\eeq}{\end{equation}}
\newcommand{\beqa}{\begin{eqnarray}}
\newcommand{\eeqa}{\end{eqnarray}}
\newcommand{\beqas}{\begin{eqnarray*}}
\newcommand{\eeqas}{\end{eqnarray*}}
\def\bal#1\eal{\begin{align}#1\end{align}}
\def\bals#1\eals{\begin{align*}#1\end{align*}}
\def\bsal#1\esal{\begin{small}\begin{align}#1\end{align}\end{small}}
\def\bsals#1\esals{\begin{small}\begin{align*}#1\end{align*}\end{small}}
\def\bsfal#1\esfal{\begin{small}\begin{flalign}#1\end{flalign}\end{small}}

\newcommand{\Net}{\textup{NN}}

\title{Compact Vision Transformer by Reduction of Kernel Complexity}

\author{Yancheng Wang, Yingzhen Yang \\
School of Computing and Augmented Intelligence\\
Arizona State University\\
Tempe, AZ 85281, USA \\
\texttt{\{ywan1053, yingzhen.yang\}@asu.edu} \\
}

\begin{document}

\maketitle

\begin{abstract}
Self-attention and transformer architectures have become foundational components in modern deep learning. Recent efforts have integrated transformer blocks into compact neural architectures for computer vision, giving rise to various efficient vision transformers. In this work, we introduce Transformer with Kernel Complexity Reduction, or KCR-Transformer, a compact transformer block equipped with differentiable channel selection, guided by a novel and sharp theoretical generalization bound. KCR-Transformer performs input/output channel selection in the MLP layers of transformer blocks to reduce the computational cost.
Furthermore, we provide a rigorous theoretical analysis establishing a tight generalization bound for networks equipped with KCR-Transformer blocks. Leveraging such strong theoretical results, the channel pruning by KCR-Transformer is conducted in a generalization-aware manner, ensuring that the resulting network retains a provably small generalization error.
Our KCR-Transformer is compatible with many popular and compact transformer networks, such as ViT and Swin, and it reduces the FLOPs of the vision transformers while maintaining or even improving the prediction accuracy. In the experiments, we replace all the transformer blocks in the vision transformers with KCR-Transformer blocks, leading to KCR-Transformer networks with different backbones. The resulting TCR-Transformers achieve superior performance on various computer vision tasks, achieving even better performance than the original models with even less FLOPs and parameters.
\end{abstract}

\section{Introduction}
\label{sec:introduction}

Inspired by the groundbreaking success of the Transformers~\citep{vaswani2017attention} in natural language processing, vision transformers have demonstrated promising performance on a variety of computer vision tasks~\citep{yuan2021tokens, dosovitskiy2020image, liu2021swin, zhu2020deformable, liang2021swinir} and multi-modal learning tasks~\citep{llava, flava}. However, the superior performance of vision transformers comes at the cost of substantial computational overhead~\citep{dosovitskiy2020image, touvron2021training}, posing challenges for deployment in resource-constrained environments. To reduce the computational costs of the vision transformers, various model compression methods have been developed, including knowledge distillation~\citep{zhao2023cumulative, yang2022vitkd}, quantization~\citep{li2022q, LinZSLZ22, li2023repq, liu2021post}, neural architecture search~\citep{gong2022nasvit, SuYXZWQZWX22}, and pruning~\citep{SViTE, ViT-Slim, SAViT, UVC, WDPruning, rao2021dynamicvit, SPViT, VTC-LFC, ToMe, bonnaerens2023learned, kim2024token}. The pruning methods, which typically involve a pruning stage to remove redundant parameters and a fine-tuning stage to recover performance, have been shown to be particularly effective due to the substantial parameter redundancy in ViT models~\citep{SViTE, ViT-Slim, SAViT, UVC, WDPruning, rao2021dynamicvit, SPViT, VTC-LFC, ToMe, bonnaerens2023learned, kim2024token}.
Despite substantial progress, the conventional model compression methods primarily focus on identifying optimal compression strategies through direct performance–efficiency trade-offs, guided by empirical heuristics rather than principled theoretical foundations.

More recently, inspired by advances in the theoretical understanding of deep neural networks (DNNs) through the lens of kernel methods, such as the Neural Tangent Kernel (NTK)~\citep{JacotHG18-NTK}, the kernel-based compression method has emerged as a principled alternative.
The kernel-based methods aim to preserve the generalization capability of the compressed model by ensuring that the compressed model retains the training dynamics, convergence behavior, and inductive biases of the original network through spectral alignment of the NTK~\citep{WeiGCH23, ChenGW21, MokNKHY22, WangL023, RachwanZCGAG22}. For instance, NTK-SAP~\citep{WangL023} and Early-Lottery~\citep{RachwanZCGAG22} leverage spectral preservation of the NTK during pruning to maintain the eigenspectrum, thereby preserving the generalization characteristics of the NTK of the original DNNs.
Despite recent progress, a substantial gap persists between theory and practice in enhancing the generalization capability of compressed DNNs through NTK-based kernel learning. Existing theoretical frameworks, particularly those grounded in NTK analyses, are predominantly limited to over-parameterized DNNs~\citep{CaoG19a-sgd-wide-dnns,AroraDHLW19-fine-grained-two-layer,Ghorbani2021-linearized-two-layer-nn}, rendering them unsuited for modern architectures such as vision transformers with finite-width and diverse network architecture. Moreover, the linearized nature of the NTK regime inherently fails to model the dynamically evolving kernel characteristic of realistic training dynamics, thereby limiting its applicability to compressed models utilizing large-scale training for compelling performance on real-world tasks~\citep{Nichani0L22-escape-ntk,DamianLS22-nn-representation-learning,TakakuraS24-mean-field-two-layer}.
To address this issue, we first provide a theoretical analysis that establishes a sharp generalization bound in Theorem~\ref{theorem:optimization-linear-kernel}. Both an upper and lower bound of the expected loss, referred to as KCR upper and lower bounds, are established based on the training loss and the kernel complexity (KC) of the kernel gram matrix computed over the training data, and the KC measures the complexity of the dynamically evolving kernel formed by the DNN during the training process.
Since the training loss is usually optimized to a small value by training the DNN, the KCR upper and lower bounds in Theorem~\ref{theorem:optimization-linear-kernel} can be tight and close to the expected loss if the KC is small. However, the computation of the KC involves the costly computation of the eigenvalues of a potentially large-scale gram matrix. To mitigate this issue, we introduce an approximate TNN through the Nyström method~\citep{KumarMT12}, and the KC is empirically reduced by the reduction of the approximate TNN. The approximate TNN is computed efficiently as a regularization term to the regular cross-entropy training loss. Since the approximate TNN is separable, it can be optimized by the standard SGD-based optimization algorithms. 
Based on the reduction of the KC, we propose a novel vision transformer termed the Transformer with Kernel Complexity Reduction, or KCR-Transformer. The training of the KCR-Transformer involves a search stage and a retrain stage.
The channel selection in the input/output features of the MLP in the transformer block of the KCR-Transformer is performed in the search stage to obtain a compressed network architecture with reduced computation costs. To guarantee the generalization capability of the compressed model, the compressed network is retrained with the approximate TNN as a regularization term to reduce the KC, leading to enhanced prediction accuracy.

\subsection{Contributions}
The contributions of this paper are presented as follows.

First, we present a compact transformer block termed Transformer with Kernel Complexity Reduction, or KCR-Transformer. By selecting the channels in the input and output features of the MLP layers via channel pruning, KCR-Transformer blocks effectively reduce the computational costs of the vision transformer. KCR-Transformer blocks can be used to replace all the transformer blocks in many popular vision transformers, rendering compact vision transformers with comparable or even better performance. The effectiveness of KCR-Transformer is evidenced by replacing all the transformer blocks with KCR-Transformer blocks in popular vision transformers such as ViT~\citep{dosovitskiy2020image}, and Swin~\citep{liu2021swin}. Experimental results show that KCR-Transformer not only reduces the number of parameters and FLOPs but also outperforms the original models on tasks including image classification, object detection, and instance segmentation.

Second, we provide a theoretical analysis showing a sharp generalization bound for the KCR-Transformer network. With such strong theoretical results in Theorem~\ref{theorem:optimization-linear-kernel}, the channel pruning by KCR-Transformer is performed in a \textit{generalization-aware} manner. That is, the channel pruning of KCR provably keeps a small generalization error bound for the DNN with KCR-Transformer blocks, effectively guaranteeing the generalization capability of the DNN after channel pruning. This goal is achieved through the reduction of the kernel complexity (KC) of the DNN with KCR-Transformer blocks. Since the KC involves the computation of the eigenvalues of a potentially large-scale gram matrix, we introduce an approximate truncated nuclear norm (TNN) through the Nyström method~\citep{KumarMT12}, which is computed efficiently as a regularization term to the regular cross-entropy training loss and separable, so that it can be optimized by the standard SGD-based optimization algorithms. The reduction of the approximate TNN effectively reduces the KC, leading to the superior prediction accuracy of the compressed vision transformers by KCR-Transformer.

This paper is organized as follows. The related works in efficient vision transformers and compression of vision transformers are discussed in Section~\ref{sec:related-works}. The formulation of KCR-Transformer with our theoretical results is detailed in Section~\ref{sec:formulation}. The effectiveness of KCR-Transformer is demonstrated in Section~\ref{sec:experiments} for image classification, semantic segmentation, and object detection tasks.
Throughout this paper, we use $a \lsim b$ to denote $a \le Cb$ if there exists such a positive constant $C$, and $[n]$ denotes all the natural numbers between $1$ and $n$
inclusively.


\section{Related Works}\label{sec:related-works}
\subsection{Efficient Vision Transformers and Compression of Vision Transformers}
Vision transformer architectures have become a compelling alternative to traditional convolutional neural networks (CNNs) for a variety of vision tasks, including image classification~\citep{dosovitskiy2021an, liu2021swin}, object detection~\citep{carion2020end, zhu2020deformable}, and image restoration~\citep{liang2021swinir, wang2022adaptive}. Despite their success, these models often incur substantial computational overhead due to the quadratic complexity of point-wise attention mechanisms and the heavy reliance on multi-layer perceptrons (MLPs). To mitigate these issues, various strategies have been proposed. Sparse attention mechanisms have been introduced to reduce computational demands~\citep{zhu2020deformable, yuan2021tokens, papa2024survey}, while other efforts integrate convolutional operations into the transformer architecture~\citep{cai2022efficientvit, mobilevit, yuan2021tokens, bravo2024cvtstego}. Additional gains in efficiency have been realized through Neural Architecture Search~\citep{chen2021autoformer, gong2022nasvit, wei2024auto} and Knowledge Distillation~\citep{graham2021levit, radosavovic2020designing, gong2022nasvit, yang2024vitkd}, which aim to maintain accuracy with reduced computational resources.
To further compress vision transformers, pruning techniques have been extensively explored and generally fall into three categories. Channel pruning aims to eliminate redundant attention heads and channels, as demonstrated in works such as SViTE~\citep{SViTE}, ViT-Slim~\citep{ViT-Slim}, SAViT~\citep{SAViT}, and LPViT~\citep{xu2024lpvit}, often employing structured or jointly optimized strategies. Block pruning reduces the depth and width of models by removing entire transformer blocks, with methods like UVC~\citep{UVC}, WDPruning~\citep{WDPruning}, and UPDP~\citep{liu2024updp} leveraging dynamic block skipping and shallow classifier integration. Token pruning techniques, including DynamicViT~\citep{rao2021dynamicvit}, SPViT~\citep{SPViT}, ToMe~\citep{ToMe}, and VTC-LFC~\citep{VTC-LFC}, improve efficiency by adaptively discarding, merging, or filtering less informative tokens~\citep{liu2024revisiting, mao2025prune}.

\subsection{Related Works about Kernel Methods for Deep Neural Networks (DNNs)}

Kernel methods have offered a principled view for analyzing the training dynamics, generalization properties, and architectural components fo DNNs. One of the most prominent lines of work centers around the neural tangent kernel (NTK)~\citep{jacot2018neural}, which provides a theoretical framework to understand convergence and generalization in the infinite-width limit, where neural networks behave like kernel methods under gradient descent. Subsequent studies have extended NTK theory to better capture realistic scenarios, including finite-width settings~\citep{seleznova2022analyzing}, deep narrow networks~\citep{lee2022neural}, and the empirical evolution of the NTK during training~\citep{fort2020deep}.
Following these, researchers have also studied the limitations of purely kernel-based theories in capturing the full expressivity of deep neural networks~\citep{woodworth2020kernel, barrett2021implicit}.
Recent works have examined kernel-based interpretations of feature learning and generalization, revealing how hierarchical or implicit kernel structures emerge within deep models~\citep{montavon2011kernel, belkin2018understand, xiao2020disentangling, canatar2022kernel, deng2022neuralef}. Building on these theoretical foundations, recent efforts propose reproducing kernel Hilbert space (RKHS) representations and operator-theoretic formulations as a basis for deep learning~\citep{hashimoto2023deep}, and develop hierarchical kernels tailored for representation learning~\citep{huang2023hierarchical}.
Beyond theoretical analysis, the study of kernels has also inspired reinterpretations and enhancements of transformer architectures. Several studies reframe self-attention as a kernel operation~\citep{song2021implicit, chen2023primal}. Others leverage spectral or integral transforms grounded in kernel theory~\citep{nguyen2022fourierformer, NguyenNHBBO23}. Positional encoding has also benefited from this perspective, with kernelized relative embeddings proposed for improved sequence extrapolation~\citep{chi2022kerple}. Efficient attention variant, Performer~\citep{ChoromanskiLDSG21}, exploits kernel approximations to achieve linear complexity while maintaining expressiveness. Additionally, kernel-based models have been used to improve calibration in transformers via sparse Gaussian processes~\citep{chen2023calibrating}.

\noindent\textbf{Kernel-Based Model Compression Methods.} Building upon these insights, kernel-based methods, especially those centered on the NTK, provide a complementary theoretical framework for analyzing and guiding model compression. The NTK-Comp framework~\citep{GuDZXPQL22} investigates pruning in wide multilayer perceptrons under Gaussian input assumptions and introduces quantization techniques that preserve the NTK spectrum within linear layers. MLP-Fusion~\citep{WeiGCH23} advances transformer compression by clustering neurons to jointly approximate functional outputs and NTK similarity, achieving strong performance on large language models. NTK-based metrics have also enabled training-free architecture search~\citep{ChenGW21}, and facilitated early-stage performance prediction in neural architecture search~\citep{MokNKHY22}, though their predictive power may diminish in regimes dominated by highly non-linear dynamics. In addition, methods, such as NTK-SAP~\citep{WangL023} and Early-Lottery~\citep{RachwanZCGAG22}, further highlight the importance of preserving NTK spectral properties during pruning, emphasizing spectral alignment as critical for maintaining stable training dynamics. Nonetheless, the core limitation of existing NTK-based compression methods lies in their dependence on static or “lazy” training regimes, limiting their applicability to models with dynamically evolving representations.




\section{Formulation}
\label{sec:formulation}

In this section, we present the channel selection for attention outputs in our KCR-Transformer, which prunes the channels of the MLP features so as to reduce the FLOPs of the transformer block. We then present the novel upper bound for the Kernel Complexity, and introduce the training algorithm of the network with KCR-Transformer for minimizing the upper bound.

\subsection{Channel Selection for Attention Outputs}
\label{sec:channel-selection-attention-output}
After applying the multi-head self-attention on the input patch embedding $x$, we obtain the attention outputs $z\in \mathbb{R}^{N\times D}$. Our KCR-transformer block then applies MLP layers to the attention outputs. The MLP layers in visual transformer blocks are usually computationally expensive. To improve the efficiency of the KCR-transformer, we propose to prune the channels in the attention outputs so that the computation cost of MLP layers can be reduced. Similar to the channel selection for attention weights, we maintain a decision mask $g_i \in \{0,1\}^{D}$, where $g_i=1$ indicates that the $i$-th channel is selected, and $0$ otherwise. Thus, the informative channels can be selected by multiplying $g$ by each row of the attention output. To optimize the binary decision mask with gradient descent, we replace $g$ with Gumbel Softmax weights in the continuous domain, which is computed by
$\hat g_i = \sigma \Bigl( \frac{ \alpha_i + \epsilon_i^{(1)} - \epsilon_i^{(2)} }{\tau} \Bigr)$,
where $\epsilon_i^{(1)}$ and $\epsilon_i^{(2)}$ are Gumbel noise. $\tau$ is the temperature. $\alpha \in \mathbb{R}^{D}$ is the sampling parameter. We define $\alpha$ as the architecture parameters of the KCR-Transformer block that can be optimized by gradient descent during the differentiable search process. By gradually decreasing the temperature $\tau$ in the search process, $\alpha_i$ will be optimized such that $g_i$ will approach $1$ or $0$. Note that since the MLP layers in vision transformers have the same input and output dimensions, so we multiply the decision mask $g$ with both the input and output features of the MLP layers. After the search is finished, we apply the gather operation on the attention outputs from the selected channels. The dimension of the input and output features of the MLP layers are then changed to $\tilde D = \sum_{i=1}^{D}g_i$.
In a similar manner, we can have KCR-X, where $X$ stands for a visual transformer.

\subsection{Generalization Bound by Approximate Kernel Complexity Loss and Its Optimization}
\label{sec:IB-optimization}

Given the training data $\set{\bX_i,y_i}_{i=1}^n$ where $\bX_i$ is the $i$-th input training feature, $y_i \in [C]$ is the corresponding class label and $C$ is the number of classes. We denote the label matrix as $\bY \in \RR^{n\times C}$ where the $i$-th row is the one-hot vector corresponding to the class label $y_i$ for all $i \in [n]$. Let $\bF\in \RR^{n\times d}$ be the features extracted on the entire training data set, where $g(\cdot) \in \RR^d$ denotes the mapping function of a DNN, such as ViT~\citep{dosovitskiy2020image}, and $d$ is the output dimension of the DNN before the final softmax layer for classification.
We remark that almost all the DNNs use a linear layer to generate the output of the network for discriminative learning tasks, so that the mapping function of a DNN can be formulated as $g(\cdot) = g(\cW,\cdot) = F(\cW_2,\cdot) \cW_1$, where $\cW_1 \in \RR^{m \times C}$ contains the weights in the final linear layer of the DNN where $m$ is the hidden dimension, $F(\cW_2,\cdot) \in \RR^m$ represents the feature extraction backbone of the network before the final linear layer, and $\cW_2$ are the weights of the backbone $F$, with $\cW = \set{\cW_1,\cW_2}$. The DNN is denoted as
$\Net_{\cW}(\cdot)$. It is noted that such formulation does not impose any limitation on the feature backbone $F$ so as to admit a broad class of DNNs with various architectures for real-world vision discriminative tasks. We define a positive definite kernel for the DNN as
\bal\label{eq:K-backbone=kernel}
K(\bx,\bx') = F(\bx)^{\top} F(\bx'), \quad \forall \bx,\bx' \in \cX,
\eal
where $\cX$ is the input domain of the DNN. Let $\bF_i = F(\cW_2,\bX_i)$ be the learned representation for the $i$-th training data and $\bF \in \RR^{n \times d}$ are the matrix of all the learned representations over the training data. Then the gram matrix $\bK \in \RR^{n \times n}$ of the kernel $K$ over the training data is calculated by $\bK = \bF^{\top}\bF \in \RR^{n \times n}$, and the eigenvalues of $\bK_n \defeq \bK/n$ are $\hlambda_1 \ge \hlambda_2 \ldots \ge \hlambda_{r_0} \ge \hlambda_{r_0+1} = \ldots  =  \hlambda_n = 0$ with $r_0 = \min\set{n,m}$, since $\rank(\bK) \le r_0$.

Suppose the input feature $\bx$ and its class label $y$ follow an unknown joint distribution $P$ over $\cX \times [C]$, then
the expected risk of the DNN is defined as $L_{\cD}(\Net_{\cW}) = \Expect{(\bx,y) \in P}{\ltwonorm{\Net_{\cW}(\bx)-y}^2}$, which also represents the generalization error of the DNN. The following theorem, based on the local complexity of the function class of the DNN feature extraction backbones and rooted in the well-established local Rademacher complexity literature~\citep{bartlett2005,koltchinskii2006,Mendelson02-geometric-kernel-machine}, gives a sharp bound for the generalization error of the DNN. We define the (empirical) kernel complexity (KC) of the kernel $K$ over the training data as $\textup{KC}(\bK) \defeq \min\limits_{h \in [0,r_0]} \pth{\frac hn + \sqrt{\frac 1n \sum\limits_{i=h+1}^{r_0} \hlambda_i}}$.
\begin{theorem}
\label{theorem:optimization-linear-kernel}
For every $x > 0$, with probability at least $1-\exp(-x)$, after the $t$-th iteration of gradient descent for all $t \ge 1$ on $\cW_1$, we have
\bal\label{eq:optimization-linear-kernel-generalization}
\underbrace{\fnorm{\pth{\bI_n - \eta \bK_n }^t \bY}^2 - \textup{KC}(\bK) - \frac{x}{n}}_{\textup{KCR Lower Bound}} \lsim L_{\cD}(\Net_{\cW}) &\lsim \underbrace{\fnorm{\pth{\bI_n - \eta \bK_n }^t \bY}^2 + \textup{KC}(\bK) + \frac{x}{n}}_{\textup{KCR Upper Bound}}.
\eal
\end{theorem}
The proof of this theorem is deferred to Section~\ref{sec:proofs} of the supplementary.
The training loss $\fnorm{\pth{\bI_n - \eta \bK_n }^t \bY}^2$ is usually optimized to a small value by training the network $\Net_{\cW}$, so it can be observed from (\ref{eq:optimization-linear-kernel-generalization}) the KCR upper and lower bounds for $L_{\cD}(\Net_{\cW})$ can be tight and close to $L_{\cD}(\Net_{\cW})$ if the kernel complexity  $\textup{KC}(\bK)$ is small. In order to optimize the kernel complexity, we introduce
the truncated nuclear norm  (TNN) of $\bK$, which is denoted as $\norm{\bK}{r} \defeq \sum_{i=r+1}^d \hlambda_i$ where $r \in [0\colon r_0]$, with $r_0 = \gamma_0 \min\set{n,m}$. It can be verified that a smaller TNN $\norm{\bK}{r}$ leads to a smaller KC $\textup{KC}(\bK)$. Since the computation of the TNN involves the computation of the eigenvalues of the potentially large-scale gram matrix $\bK$ over large-scale training data, we then describe below how to efficiently and effectively approximate the TNN. We first compute the approximate top-$r_0$ eigenvectors of $\bK_n$, $\tilde \bU^{r_0}$, by the Nyström method~\citep{KumarMT12}. Here $\bA^{(r)}$ denotes a submatrix of $\bA$ formed by its top $r$ columns.

\noindent \textbf{Efficient Computation of the Top-$r_0$ Eigenvectors of  $\bK$.} Next, we approximate the top-$r_0$ eigenvectors, $\bU^{(r_0)} \in \mathbb{R}^{n \times r_0}$, of the gram matrix $\bK$ using the Nyström method~\citep{KumarMT12}, we first sample $m$ landmark points from the training set, indexed by $\mathcal{I} \subset [n]$ with $|\mathcal{I}| = m \ll n$. Let $\bF_\mathcal{I} \in \mathbb{R}^{m \times d}$ be the features corresponding to the landmark set. We define $\bC = \bF \bF_\mathcal{I}^\top \in \mathbb{R}^{n \times m}$ as the cross-covariance matrix and $\cW = \bF_\mathcal{I} \bF_\mathcal{I}^\top \in \mathbb{R}^{m \times m}$ as the Gram matrix on the landmarks. Next, we compute the top-$r_0$ eigen-decomposition of $\cW$ as $\cW = \bQ \bLambda \bQ^\top$, where $\bQ \in \mathbb{R}^{m \times r_0}$ contains the top-$r_0$ eigenvectors and $\bLambda \in \mathbb{R}^{r_0 \times r_0}$ is the diagonal matrix of corresponding eigenvalues. The Nyström approximation of $\bK$ is then given by $\tilde{\bK} = \bC \cW^\dagger \bC^\top$, and the approximate top-$r_0$ eigenvectors are computed as $\tilde{\bU}^{(r_0)} = \bC \bQ \bLambda^{-1/2} \in \mathbb{R}^{n \times r_0}$, which serves as an efficient approximation to $\bU^{(r_0)}$ with significantly reduced computational cost. The value of $m$ is selected by cross-validation for each feature extraction backbone as detailed in Section~\ref{sec:classification}.

We let $\bU_r = \tilde{\bU}^{(r)}$, then the sum of the top-$r$ eigenvalues of $\bK_n$ is approximated by $\tr{{\bU_r}^{\top} \bK_n \bU_r}$.
Since $\tr{\bK_n} = \sum_{i=1}^n K(\bx_i,\bx_i) = \sum_{i=1}^n \hlambda_i$, and $\tr{{\bU_r}^{\top} \bK_n \bU_r} = \sum\limits_{i=1}^n ( \sum\limits_{s=1}^{r} \sum\limits_{k=1}^n \bth{\bU_r}^{\top}_{si} \bth{\bK_n}_{ik} \bth{\bU_r}_{ks}  )$, we can approximate the $\norm{\bK}{r}$ by
$\overline{\norm{\bK}{r}}  = \tr{\bK_n} - \tr{{\bU_r}^{\top} \bK_n \bU_r} $ which is separable. In particular,
\bal\label{eq:tilde-TNN-r}
\overline{\norm{\bK}{r}} = \sum_{i=1}^n K(\bx_i,\bx_i) - \sum\limits_{i=1}^n \pth{ \sum\limits_{s=1}^{r} \sum\limits_{k=1}^n \bth{\bU_r}^{\top}_{si} \bth{\bK_n}_{ik} \bth{\bU_r}_{ks} }
\eal
We remark that $\overline{\norm{\bK}{r}}$ is ready to be optimized by standard SGD algorithms because it is separable and expressed as the summation of losses on individual training points. Algorithm~\ref{Algorithm-KCR} describes the training process of a neural network with KCR-Transformer blocks where $\textup{KCR}(\cW)$ is a term in the training loss. In order to compute
$\textup{KCR}(\cW)$ before a new epoch starts, we need to update the variational distribution $Q^{(t)}$ at the end of the previous epoch. The following functions are needed for minibatch-based training with SGD, with the subscript $j$ indicating the corresponding loss on the $j$-th batch $\cB_j$:
\vspace{-.1in}
\bsals
\textup{KCR}_{j}(\cW) = \frac{1}{\abth{\cB_j}}\pth{\sum\limits_{i \in \bth{\abth{\cB_j}}}
K(\bx_i,\bx_i) - \sum\limits_{s=1}^{r} \sum\limits_{k=1}^n \bth{\bU_r}^{\top}_{si} \bth{\bK_n}_{ik} \bth{\bU_r}_{ks}
},
\esals
\vspace{-.2in}
\bal
    \mathcal{L}^{(t)}_{\text{train},j}(\cW) &= \text{CE}^{(t)}_{j} + \eta \textup{KCR}_{j}(\cW) ,\label{eq:train_loss}\\
    \text{CE}^{(t)}_{j} &=  \frac{1}{\abth{\cB_j}}\sum_{i=1}^{\abth{\cB_j}}H(X_i(\cW), Y_i).  \nonumber
\eal
Here $\text{CE}^{(t)}_{j}$ is the cross-entropy loss on batch $\cB_j$ at epoch $t$. $H(,)$ is the cross-entropy function. $\eta$ is the balance factor.
\begin{algorithm}[!htb]
\caption{Training Algorithm with the Approximate Truncated Nuclear Norm by SGD}\label{Algorithm-KCR}
{
\small
\begin{algorithmic}[1]
\STATE Initialize the weights of the network
by $\cW = \cW(0)$ through random initialization
\FOR{$t\leftarrow 1$ to $t_{\text{search}}$}
\FOR{$j \leftarrow 1$ to $J$}
\STATE Perform gradient descent with batch $\cB_j$ using the loss $\mathcal{L}^{(t)}_{\text{search},j}(\cW, \alpha) $ defined Section~\ref{sec:search}.
\ENDFOR
\ENDFOR

\FOR{$t\leftarrow 1$ to $t_{\text{train}}$}
\FOR{$j \leftarrow 1$ to $J$}
\IF{$t>t_{\text{warm}}$}
\STATE Perform gradient descent with batch $\cB_j$ using the loss $\mathcal{L}^{(t)}_{\text{train},j}(\cW) $ by Eq. (\ref{eq:train_loss}).
\ELSE
\STATE Perform gradient descent with batch $\cB_j$ using the loss $\text{CE}^{(t)}_{j}$ by Eq. (\ref{eq:train_loss}).
\ENDIF
\ENDFOR
\ENDFOR
\STATE \textbf{return} The trained weights $\cW$ of the network
\end{algorithmic}
}
\end{algorithm}

\subsection{Optimization in the Search Process}
\label{sec:search}
To obtain a compact visual transformer network with KCR-Transformer, we need to optimize both the accuracy of the network and the inference cost (FLOPs) of the network. Therefore, the differentiable inference cost of the network needs to be estimated and optimized during the search phase.
It is worthwhile to mention that we follow the standard techniques in neural architecture search \citep{tan2019mnasnet} in the searching process, including channel selection by Gumbel-Softmax and entropy minimization for architecture search. We optimize the FLOPs of the operations whose computation cost is decided by the channel selection for attention outputs in Section~\ref{sec:channel-selection-attention-output}. We estimate the FLOPs of the MLP after the channel selection on the attention outputs following the KCR-Transformer. The estimation of the FLOPs related to a single Transformer block is $\mathtt{cost}_j = l_j\cdot \pth{ 2 \tilde D^2 + \tilde D}$,
where $j$ indexes the KCR-Transformer block. $2 \tilde D^2 + \tilde D$ is the FLOPs of a layer of the MLP after the channel selection on the attention outputs, and $l_j$ denotes the number of layers in the MLP of the $j$-th KCR-Transformer block. As a result, we can calculate the inference cost objective of the network architecture by $\mathtt{cost} = \sum_{j=1}^{M} \mathtt{cost}_j$, where $M$ is the number of transformer blocks.
To supervise the search process, we designed a loss function incorporating cost-based regularization to enable multi-objective optimization. The overall loss function for search on each batch $\cB_j$ at epoch $t$ is formulated
by $\mathcal{L}^{(t)}_{\text{search},j}(\cW, \alpha) = \text{CE}^{(t)}_{j} + \lambda \cdot \log \mathtt{cost}(\alpha), ~\text{CE}^{(t)}_{j} =  \frac{1}{\abth{\cB_j}}
    \sum_{i=1}^{\abth{\cB_j}}H(X_i(\cW), Y_i)$,
where $\cW$ denotes the weights in the supernet. $\alpha$ is the architecture parameter. $\lambda$ is the hyperparameter that controls the magnitude of the cost term, which is selected by cross-validation. In the search phase, the search loss is optimized to perform the two types of channel selection for all the KCR-Transformer blocks. After the search phase, we use the selected channels for the attention outputs in a searched network and then perform retraining on the searched network.
\vspace{-.1in}
\section{Experimental Results}
\label{sec:experiments}
In this section, we first evaluate the performance of KCR-Transformers on the ImageNet-1k dataset for image classification in Section~\ref{sec:classification}. In Section~\ref{sec:loss_study}, we study the expected loss computed and the approximated KCR upper/lower bounds of the KCR-Transformers.
In Section~\ref{sec:segmentation} and Section~\ref{sec:object_detection}, we study the effectiveness of using KCR-Transformer as the feature extraction backbone for semantic segmentation and object detection. In Section~\ref{sec:ablation-study}, we study the effectiveness of the KCR-Transformer in reducing the KC of the networks.

\subsection{Image Classification}
\label{sec:classification}
\begin{wrapfigure}{R}{0.6\textwidth}
    \centering
    \vspace{-10pt} 
    \begin{minipage}{0.58\textwidth}
    \small
    \resizebox{\textwidth}{!}{
        \begin{tabular}{lrrr}
            \toprule[1.5pt]
            Model  & \# Params & FLOPs & Top-1 \\
            \midrule[1pt]
            T2T~\citep{yuan2021tokens}  & 4.3 M &1.1 G & 71.7 \\
            DeIT~\citep{touvron2021training}  & 5.7 M & 1.2 G & 72.2 \\
            CrossViT~\citep{chen2021crossvit}  & 6.9 M & 1.6 G & 73.4 \\
            DeIT~\citep{touvron2021training}  & 10 M &2.2 G & 76.6 \\
            T2T~\citep{yuan2021tokens} & 6.9 M & 1.8 G& 76.5 \\
            PiT~\citep{heo2021rethinking}  & 10.6 M & 1.4 G & 78.1 \\
            Mobile-Former~\citep{chen2021mobile} & 9.4 M  & 0.2 G & 76.7 \\
            EViT~\citep{liu2023efficientvit}  & 12.4 M & 0.5 G & 77.1 \\
            TinyViT~\citep{wu2022tinyvit}  & 5.4 M & 1.3 G & 79.1 \\
            DeIT~\citep{touvron2021training}  & 22 M & 4.6 G & 79.8 \\
            EfficientFormer~\citep{li2022efficientformer}  & 12.3 M & 1.3 G & 79.2 \\
            VTC-LFC~\citep{VTC-LFC}& 5.0 M & 1.3 G& 78.0 \\
            SPViT~\citep{SPViT}& 4.9 M & 1.2 G& 77.8\\
            \midrule
            ViT-S~\citep{dosovitskiy2020image}  & 22.1 M & 4.3 G &81.2\\
            NTK-SAP-ViT-S~\citep{WangL023}  & 20.3 M & 3.9 G &80.9\\
            \textbf{KCR-ViT-S (Ours)}~  &  19.8 M & 3.8 G& \textbf{82.2} \\
            ViT-B~\citep{dosovitskiy2020image}  & 86.5 M & 17.6 G &83.7\\
            NTK-SAP-ViT-B~\citep{WangL023}  & 71.8 M & 15.6 G &83.5\\
            \textbf{KCR-ViT-B (Ours)}~  & 69.5 M & 14.5 G &\textbf{84.6}\\
            Swin-T~\citep{liu2021swin}  & 29.0 M & 4.5 G &81.3\\
            NTK-SAP-Swin-T~\citep{WangL023}  & 25.5 M & 4.2 G &81.2\\
            \textbf{KCR-Swin-T (Ours)}~  & 24.6 M & 3.9 G &\textbf{82.4}\\
            Swin-B~\citep{liu2021swin}  & 88.0 M & 15.4 G &83.5\\
            NTK-SAP-Swin-B~\citep{WangL023}  & 72.6 M & 13.2 G &83.2\\
            \textbf{KCR-Swin-B (Ours)}~  & 70.2 M& 12.6 G& \textbf{84.7}\\
            \bottomrule[1.5pt]
        \end{tabular}
    }
    \captionof{table}{Comparisons with baseline methods on ImageNet-1k validation set.}
    \label{tab:imagenet_results}
    \end{minipage}
\end{wrapfigure}
\textbf{Implementation details about Search/Retraining.}
For our ImageNet classification experiments, we adopt ViT-S~\citep{dosovitskiy2020image}, ViT-B~\citep{dosovitskiy2020image}, Swin-T~\citep{liu2021swin}, and Swin-B~\citep{liu2021swin} as backbone models. Each transformer block in these architectures is substituted with a KCR-Transformer block. During the architecture search phase, we randomly sample a subset of $100$ ImageNet classes~\citep{russakovsky2015imagenet} for training. The network is optimized using the AdamW optimizer with a cosine learning rate schedule, where the initial learning rate of $0.001$ is gradually annealed to $0.0001$ over $200$ epochs. The optimizer parameters are set to $\beta_{1}=0.9$ and $\beta_{2}=0.999$. In each epoch, $70\%$ of the training samples are used for updating the model weights, while the remaining $30\%$ are dedicated to optimizing the architecture parameters of the KCR-Transformer blocks. The temperature $\tau$ is initialized at $4.5$ and decayed by a factor of $0.95$ per epoch.
Upon completion of the search phase, the final architecture is sampled from the supernet and retrained from scratch. The parameters $t_{\text{warm}}$ and $\eta$ are determined through cross-validation, following the procedure detailed in Section~\ref{sec:search-retrain} of the supplementary material. Empirical results indicate that setting $t_{\text{warm}}=90$ and $\eta=1$ yields the best performance across all KCR model variants.

During the retraining phase, the searched network is trained on the training set of ImageNet-1K using the AdamW optimizer with $\beta_1 = 0.9$ and $\beta_2 = 0.999$. Training is conducted for $300$ epochs, with the first $90$ epochs serving as a warm-up period during which only the cross-entropy loss is optimized. After this stage, the full training objective, comprising both the cross-entropy loss and the upper bound of the kernel complexity, is used. All experiments are conducted on four NVIDIA V100 GPUs with an effective batch size of $512$. Following standard practice~\citep{cai2022efficientvit}, we apply widely adopted data augmentation techniques during training, including random scaling, random horizontal flipping, and random cropping. The weight decay is set to $0.01$. The learning rate is linearly increased from $0.0002$ to $0.002$ over the first five epochs, then gradually annealed back to $0.0002$ using a cosine decay schedule over the remaining epochs. Inference is performed using the exponential moving average (EMA) of model weights.

\textbf{Tuning Hyper-Parameters by Cross-Validation.}
To decide the best balancing factor $\lambda$ for the overall search loss, $\eta$ for the overall training loss, and the number of warm-up epochs $t_{\text{warm}}$, we perform 5-fold cross-validation on $10\%$ of the training data. The value of $\lambda$ is selected from $0.1$ to $0.5$ with a step size of $0.05$. The value of $\eta$ is selected from $\{0.1, 0.5, 1, 5, 10, 50, 100\}$. The value of $t_{\text{warm}}$ is selected from $\{0.1t_{\text{train}}$, $0.2t_{\text{train}}$, $0.3t_{\text{train}}$, $0.4t_{\text{train}}$, $0.5t_{\text{train}}$, $0.6t_{\text{train}}\}$, where $t_{\text{train}}=300$ is the total number of training epochs.  We select the values of $\eta$, $\lambda$, and $t_{\text{warm}}$ that lead to the smallest validation loss. It is revealed that $t_{\text{warm}} = 90$ is chosen for all the vision transformers in our experiments.
In addition, we search for the optimal values of feature rank $r$. Let $r = \lceil \gamma \min(n, d) \rceil$, where $\gamma$ is the rank ratio. We select the values of $\gamma$ and $\eta$ by performing 5-fold cross-validation on 20\% of the training data in each dataset. The value of $\gamma$ is selected from $\set{0.05, 0.10, 0.15, 0.20, 0.25, 0.30, 0.35, 0.4, 0.45, 0.50}$. We denote $\gamma_0 = 0.5$ as the maximum value of all the candidate $\gamma$. The value of $m$ is selected from $\set{10000, 50000, 100000, 150000, 200000}$ by 5-fold cross-validation as well. It is revealed that $100000$ is chosen for all the vision transformers in our experiments.

It can be observed from Table~\ref{tab:imagenet_results} that models with KCR-Transformer always enjoy fewer FLOPs than their original visual transformer and better accuracy. For example, KCR-Swin-B achieves a
$1.1\%$ improvement in Top-1 accuracy while reducing FLOPs from $15.4$ G to $12.6$ G compared to the original Swin-B.

\subsection{Study on the Expected Loss of KCR-Transformers }
\label{sec:loss_study}
\begin{figure}[!htb]
\begin{center}
\includegraphics[width=1\textwidth]{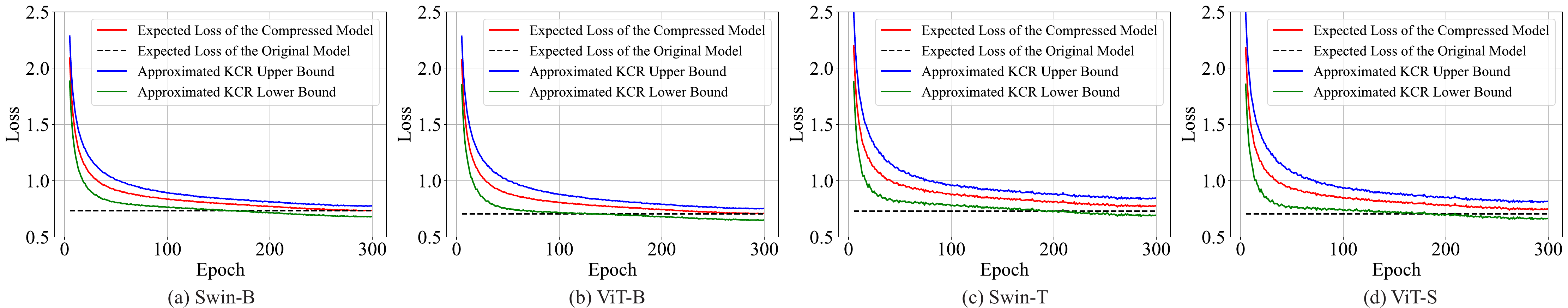}
\caption{Illustration of the expected loss and the approximated KCL upper/lower bounds over different training epochs for ViT-S, ViT-B, Swin-T, and Swin-B compressed by KCR.}
\label{fig:kcl_loss}
\end{center}
\end{figure}

Figure~\ref{fig:kcl_loss} illustrates the expected loss computed over the training and validation sets of the ImageNet-1K and the approximated KCR upper/lower bounds computed at different training epochs for ViT-S, ViT-B, Swin-T, and Swin-B trained on the ImageNet-1K dataset. We note that the approximated KCR upper/lower bounds are the KCR upper/lower bounds with the KC replaced by the approximate KC, A-KC, which is defined as $\textup{A-KC}(\bK) \defeq \min\limits_{h \in [0,r_0]} \pth{\frac hn + \sqrt{\frac 1n \overline{\norm{\bK}{h}}}}$. Since the approximate TNN $\overline{\norm{\bK}{h}}$ is expected to be close to the TNN $\norm{\bK}{h}$ for each $h \in [0\colon r_0]$, the approximate KC $\textup{A-KC}(\bK)$ is also expected to be close to the KC $\textup{KC}(\bK)$.
It can be observed that the approximated KCR upper/lower bounds are tightly correlated to the expected loss, revealing the tightness of the upper/lower bounds for the generalization error of the DNNs with KCR-Transformer blocks.

\subsection{Instance Segmentation}
\label{sec:segmentation}
In this section, we evaluate the performance of KCR for the task of instance segmentation on the ADE20K\citep{ADE20K} dataset, which covers a broad range of $150$ semantic categories. ADE20K has $25000$ images in total, with $20000$ for training, $2000$ for validation, and another $3000$ for testing. We adopt UperNet~\citep{XiaoLZJS18} as the segmentation framework with our KCR-Swin-B as the feature extraction backbone. We include Swin-B~\citep{liu2021swin} and SETR~\citep{ZhengLZZLWFFXT021} as baselines for comparisons. We follow the training and evaluation protocol in~\citep{liu2021swin}, where both our model and the baselines are trained on the training split and evaluated on the validation split of the dataset. All models are optimized using AdamW for a total of $160000$ iterations with a batch size of $16$, an initial learning rate of $6\times 10^{-5}$, and a weight decay of $0.01$. The learning rate follows a linear decay schedule after a warm-up phase of $1500$ iterations. To enhance generalization, we employ data augmentation techniques including random horizontal flipping, random rescaling with a ratio range of $[0.5, 2.0]$, and random photometric distortions. Stochastic depth regularization is applied with a drop rate of $0.2$. For inference, we use multi-scale testing with scale factors varying from $0.5$ to $1.75$.
Following~\citep{liu2021swin}, we evaluate performance using the mean Intersection over Union (mIoU), computed as the average IoU across all semantic categories. For each class, the IoU is defined as the ratio between the number of correctly predicted pixels and the total number of pixels belonging to either the prediction or the ground truth. The results are presented in Table~\ref{tab:seg_results}.
\begin{table}[!htbp]
 \small
 \centering
  \resizebox{0.6\linewidth}{!}{
\begin{tabular}{lcc}
\toprule
Segmentation Framework & Feature Backbone           & Val mIoU   \\
\midrule
SETR~\citep{ZhengLZZLWFFXT021} & ViT-L          &   50.3 \\
UperNet~\citep{XiaoLZJS18} & Swin-B           & 51.6   \\
UperNet & KCR-Swin-B (Ours)          &  52.4  \\
\bottomrule
 \end{tabular}
 }
 \caption{Instance Segmentation Results on ADE20K.}
 \label{tab:seg_results}
\end{table}

\subsection{Object Detection}
\label{sec:object_detection}
We incorporate the ImageNet pre-trained KCR-Swin-T and KCR-Swin-B into the Cascade Mask R-CNN framework~\citep{CaiV21} for object detection. All models are evaluated on the MS-COCO dataset~\citep{lin2014microsoft}, which consists of $117000$ training images and $5000$ validation images. We follow the training configuration of~\citep{liu2021swin}, where each input image is resized such that the shorter side falls within $[480, 800]$ pixels while the longer side does not exceed $1333$ pixels. Training is performed using the AdamW optimizer with an initial learning rate of $0.0001$, a weight decay of $0.05$, and a batch size of $16$, for a total of $36$ epochs following the $3\times$ schedule.
In line with~\citep{CaiV21}, we report standard COCO metrics, including the box-level mean Average Precision (mAP$^{\text{box}}$) and mask-level mean Average Precision (mAP$^{\text{m}}$), as well as AP at IoU thresholds of $50$ and $75$ in Table~\ref{tab:object_detection}. These metrics provide a comprehensive evaluation of both object localization and segmentation performance.
\begin{table}[!htbp]
 \small
 \centering
  \resizebox{0.9\linewidth}{!}{
\begin{tabular}{lccccccc}
\toprule
Detection Framework & Feature Backbone           & mAP$^{box}$ & AP$^{b}_{50}$  & AP$^{b}_{75}$  & mAP$^{m}$ & AP$^{m}_{50}$  & AP$^{m}_{75}$   \\
\midrule
Mask R-CNN& Swin-T        & 50.5               & 69.3                           & 54.9                            & 43.7                & 66.6                             & 47.1                              \\
Mask R-CNN& Swin-B        & 51.9               & 70.9                           & 56.5                          & 45.0                & 68.4                             &48.7                              \\
Mask R-CNN&  KCR-Swin-T (Ours) & 50.9               & 69.7                           & 55.3                            & 44.0                & 67.1                             & 47.6                              \\
Mask R-CNN&  KCR-Swin-B (Ours) & 52.5               & 71.4                           & 56.8                          & 45.6                & 68.9                             &49.1                              \\
\bottomrule
 \end{tabular}
 }
 \caption{Instance Segmentation Results on COCO.}
 \label{tab:object_detection}
\end{table}


\subsection{Ablation Study on the Effects of KCR-Transformer in Reducing the KCL}
\label{sec:ablation-study}

We study how KCR helps in reducing the KCL with three different vision transformers, including ViT-S, ViT-B, Swin-T, and Swin-B. We compare the performance of the vanilla vision transformers and the corresponding KCR-Tranformer models. It is observed in Table~\ref{tab:ablation-IB-loss} that KCR models exhibit significantly reduced KCL compared to the baseline models with even less parameter size and FLOPs, demonstrating enhanced generalization of the image classification task with better top-1 classification accuracy.
\begin{table}[!htbp]
 \small
 \centering
 \resizebox{0.7\columnwidth}{!}{
 \begin{tabular}{lcccc}
  \toprule
  Model & \# Params & FLOPs & Top-1 & KCL\\
  \midrule
            ViT-S~\citep{dosovitskiy2020image}  & 22.1 M & 4.3 G &81.2 &4.12\\
            \textbf{KCR-ViT-S (Ours)}~  &  19.8 M & 3.8 G& \textbf{82.2} & \textbf{0.65}\\
            ViT-B~\citep{dosovitskiy2020image}  & 86.5 M & 17.6 G &83.7 &4.35\\
            \textbf{KCR-ViT-B (Ours)}~  & 69.5 M & 14.5 G &\textbf{84.6} & \textbf{0.52}\\
            Swin-T~\citep{liu2021swin}  & 29.0 M & 4.5 G &81.3 &3.42\\
            \textbf{KCR-Swin-T (Ours)}~  & 24.6 M & 3.9 G &\textbf{82.4} &\textbf{0.44}\\
            Swin-B~\citep{liu2021swin}  & 88.0 M & 15.4 G &83.5 &3.21\\
            \textbf{KCR-Swin-B (Ours)}~  & 70.2 M& 12.6 G& \textbf{84.7} &\textbf{0.41}\\
  \bottomrule
 \end{tabular}
 }
 \caption{Ablation Study on the Effects of KCR-Transformer in Reducing the KCL.}
 \label{tab:ablation-IB-loss}
\end{table}

\section{Conclusion}
\label{sec:conclusion}
In this paper, we propose KCR-Transformer, a novel and generalization-aware transformer block equipped with differentiable channel selection for the MLP layers in vision transformers. Guided by a novel and sharp theoretical generalization bound derived from the kernel complexity (KC) of the network, KCR-Transformer enables channel pruning in a theoretically grounded and principled manner. Our method is compatible with a wide range of vision transformer architectures and can be seamlessly integrated to replace standard transformer blocks. Extensive experiments across image classification, object detection, and semantic segmentation demonstrate that KCR-Transformer consistently achieves superior with fewer FLOPs and parameters, validating its effectiveness as a building block for efficient vision transformers.



\bibliography{ref}
\bibliographystyle{iclr2025_conference}

\appendix

\section{Implementation Details}

\subsection{Settings for Searching and Retraining}
\label{sec:search-retrain}

\textbf{More Details about Retraining and Searching with KCR-Transformer.} In the retraining phase, the searched network is trained on the ImageNet-1K dataset. We employ the AdamW optimizer with hyperparameters $\beta_{1} = 0.9$ and $\beta_{2} = 0.999$. The retraining is conducted for $300$ epochs, with the first $90$ epochs dedicated to a warm-up stage that optimizes only the cross-entropy loss. After the warm-up, the full training loss is used.
Training is performed on $4$ NVIDIA V100 GPUs with an effective batch size of $512$ images. Following standard practice~\citep{cai2022efficientvit}, we adopt common data augmentation techniques, including random scaling, random horizontal flipping, and random cropping. The weight decay is set to $0.01$. The learning rate is linearly increased from $0.0002$ to $0.002$ over the first $5$ epochs, and then annealed back to $0.0002$ using a cosine decay schedule over the remaining epochs.
Inference is conducted using the exponential moving average (EMA) of the model weights. Unlike previous works such as LeViT~\citep{graham2021levit} and NASViT~\citep{gong2022nasvit}, we do not employ knowledge distillation during training. All comparisons are made against baseline models trained without knowledge distillation to ensure a fair evaluation.

Ask ChatGPT

\section{Proof of Theorem~\ref{theorem:optimization-linear-kernel}}
\label{sec:proofs}
\begin{proof}[\textup{\bf Proof of Theorem~\ref{theorem:optimization-linear-kernel}}]
It can be verified that at the $t$-th iteration of gradient descent for $t \ge 1$, we have
\bal\label{eq:optimization-linear-kernel-seg1}
\cW_1^{(t)} = \cW_1^{(t-1)} - \frac {\eta}{n} \bF^{\top}  \pth{\bF \cW_1^{(t-1)} - \bY}.
\eal
It follows by (\ref{eq:optimization-linear-kernel-seg1}) that
\bal\label{eq:optimization-linear-kernel-seg2}
\bF \cW_1^{(t)} &=  \bF \cW_1^{(t-1)} - \eta \bK_n \pth{\bF \cW_1^{(t-1)} - \bY} \nonumber \\
&=\bF \cW_1^{(t-1)} - \eta \bK_n \pth{\bF \cW_1^{(t-1)} - \bY},
\eal
where $\bK_n =  1/n \cdot \bF \bF^{\top} = 1/n \cdot \bK$.

We define $\bF(\cW,t) \defeq \bF \cW_1^{(t)}$, then it follows by (\ref{eq:optimization-linear-kernel-seg2}) that
\bals
\bF(\cW,t) - \bY = \pth{\bI_n - \eta \bK_n } \pth{\bF(\cW,t-1) - \bY},
\eals
which indicates that
\bals
\bF(\cW,t) - \bY &= \pth{\bI_n - \eta \bK_n }^t \pth{\bF(\cW,0) - \bY} \\
&= - \pth{\bI_n - \eta \bK_n }^t \bY,
\eals
since $\cW_1^{(0)} = \bzero$.
It follows from the above recursion that
\bal\label{eq:optimization-linear-kernel-full-loss}
\fnorm{\bF(\cW,t) - \bY} =  \fnorm{\pth{\bI_n - \eta \bK_n }^t \bY}.
\eal
We recall that $\textup{KC}(\bK) = \min\limits_{h \in [0,r_0]} \pth{\frac hn + \sqrt{\frac 1n
\sum\limits_{i=h+1}^{r_0} \hlambda_i}}$ is the kernel complexity.
Using the proof of ~\cite[Theorem 3.3, Corollary 6.7]{bartlett2005}, for every $x > 0$,
with probability at least $1-\exp(-x)$,
\bal\label{eq:optimization-linear-lrc-bound}
\fnorm{\bF(\cW,t) - \bY}^2 - \textup{KC}(\bK) - \frac{x}{n} \lsim L_{\cD}(\Net_{\cW}) &\lsim \fnorm{\bF(\cW,t) - \bY}^2 + \textup{KC}(\bK) + \frac{x}{n}.
\eal
Combining (\ref{eq:optimization-linear-kernel-full-loss}) and (\ref{eq:optimization-linear-lrc-bound}) proves this theorem.

\end{proof}

\end{document}